\PassOptionsToPackage{table}{xcolor}
\documentclass{article}
\usepackage{iclr2027_conference,times}
\usepackage{graphicx}
\usepackage{amsmath,amssymb}
\usepackage{pifont}
\usepackage{booktabs}
\usepackage[table]{xcolor}
\usepackage{tabularx}
\usepackage{placeins}
\usepackage{float}
\usepackage{wrapfig}
\usepackage{capt-of}
\usepackage{animate}
\usepackage[percent]{overpic}
\usepackage{hyperref}
\usepackage{url}
\definecolor{BaseRowGray}{rgb}{0.90,0.90,0.90}
\definecolor{OursRow}{rgb}{1.0,0.88,0.76}
\definecolor{TeaserOrange}{RGB}{233,113,50}
\definecolor{TeaserMagenta}{RGB}{216,110,204}
\newcommand{\revision}[1]{\textcolor{black}{#1}}
\title{WorldAlign: Decoupled 4D Reward for World-Consistent Video Generation}
\author{%
\textbf{Jing He\textsuperscript{1,3}\thanks{\mbox{This work was conducted during the author's internship at KlingAI.\quad\ding{41}\ Corresponding author.}}, Kaixin Ding\textsuperscript{3,4},
Xingye Tian\textsuperscript{3}, Guibao Shen\textsuperscript{1},}\\
\textbf{Wenhang Ge\textsuperscript{1}, Xin Tao\textsuperscript{3},
Pengfei Wan\textsuperscript{3}, Ying-Cong Chen\textsuperscript{1,2,\ding{41}}}\\[0.5em]
\textnormal{\textsuperscript{1}The Hong Kong University of Science and Technology (Guangzhou)}\\
\textnormal{\textsuperscript{2}The Hong Kong University of Science and Technology}\\
\textnormal{\textsuperscript{3}KlingAI\quad
\textsuperscript{4}The University of Hong Kong}
}
\iclrfinalcopy
\begin{document}
\maketitle
\lhead{}

\vspace{-5mm}
\noindent\begin{minipage}{\textwidth}
\centering
\begin{minipage}{\textwidth}
        \begin{overpic}[width=\linewidth]{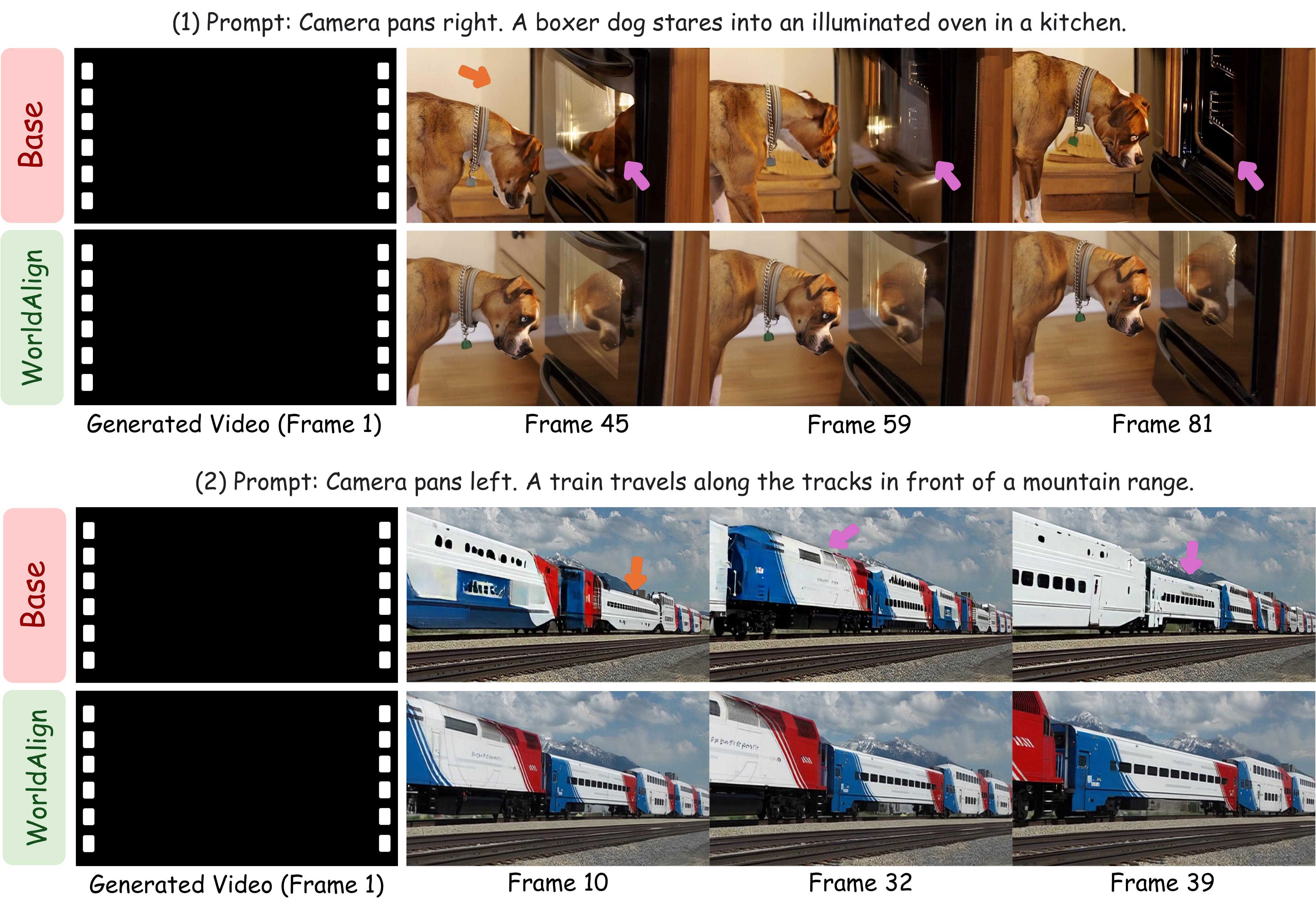}
            \put(7.52,52.935){\animategraphics[width=0.202\textwidth,autoplay,loop,every=1,poster=first]
            {8}{Figures/teaser_videos/1-base/}{0}{40}}
            \put(7.52,39.235){\animategraphics[width=0.202\textwidth,autoplay,loop,every=1,poster=first]
            {8}{Figures/teaser_videos/1-ours/}{0}{40}}
            \put(7.72,18.105){\animategraphics[width=0.202\textwidth,autoplay,loop,every=1,poster=first]
            {8}{Figures/teaser_videos/2-base/}{0}{40}}
            \put(7.72,4.355){\animategraphics[width=0.202\textwidth,autoplay,loop,every=1,poster=first]
            {8}{Figures/teaser_videos/2-ours/}{0}{40}}
        \end{overpic}
\end{minipage}
        \captionof{figure}{Comparison between WorldAlign and the base
        model (Wan2.1-I2V-14B-480P).
        The first case highlights improvements in static consistency: the base
        model loses the cabinetry (\textcolor{TeaserOrange}{orange} arrows), while the oven
        handle disappears and its door deforms before vanishing (\textcolor{TeaserMagenta}{magenta} arrows).
        The second case highlights improvements in dynamic consistency: the train cars
        abruptly change color and shape and move opposite to the locomotive's
        heading (see the cars marked by \textcolor{TeaserOrange}{same}-\textcolor{TeaserMagenta}{colored} arrows). WorldAlign instead preserves static scene structure and plausible
        dynamic evolution. 
        If the videos fail to play, \textbf{we strongly
        recommend using a multimedia-enabled PDF reader}. 
        }
        \label{fig:teaser}
\end{minipage}
\par

\begin{abstract}
\vspace{-3mm}
Faithful visual world simulation requires generated videos to maintain 4D world
consistency, encompassing both static and dynamic consistency. Static consistency requires
coherent 3D structure in static environments across viewpoints, while dynamic consistency
requires plausible subject motion and consistent appearance over time. Geometry-aware
post-training offers a promising way to improve world consistency. However, existing
methods often rely on a static-scene assumption. Even those that accommodate dynamic
scenes struggle to provide reliable static-consistency feedback, while dynamic consistency
is often overlooked or inadequately assessed. To address these limitations, we introduce
\emph{WorldAlign}, a decoupled 4D reward framework that semantically separates static
regions and dynamic subjects and provides feedback by aligning each with a world prior
suited to its assumptions. For static regions, WorldAlign aligns static geometry with a
geometric world prior through semantically guided masked reprojection, enabling more
reliable static-consistency evaluation; an auxiliary camera-motion reward discourages
nearly static solutions. For dynamic subjects, WorldAlign uses a strong vision-language
model (VLM) as a dynamic world prior and constructs a VLM-as-a-judge reward based on
sample-specific checklists that assess dynamicity, physical plausibility, shape, and
texture consistency. This decoupled design enables more effective online post-training
without requiring human preference annotations. Across two pretrained image-to-video
generators, Wan2.1 and Wan2.2, WorldAlign jointly improves static and dynamic consistency
over existing methods without suppressing overall or subject motion. These results support
decoupled world-prior alignment for more faithful visual world simulation.
Project page: \url{https://worldalign.github.io/}.
\end{abstract}

\section{Introduction}

Faithful visual world simulation requires more than frame-wise realism: the
generated world must remain coherent over time. We refer to this property as
world consistency, which comprises static and dynamic consistency. Static
consistency requires static environments to maintain coherent 3D structure
across viewpoints. Dynamic consistency requires subjects to move plausibly while
preserving their appearance over time. Such consistency is essential when video
generation models serve as world simulators for robotics
\citep{dvgwm,physisforcing}, autonomous driving
\citep{gaia1,drivedreamer,vista}, and interactive games
\citep{genie,hyworldplay,dreamxworld}. Despite substantial advances in
high-fidelity, open-domain video synthesis
\citep{sora,seedance2,klingomni,hunyuanvideo,wan}, large-scale pretrained video
generation models remain insufficient for faithful world simulation. A key
reason is that their pretraining objectives primarily optimize prediction in
RGB or latent space, leaving 3D structure and physical dynamics to be learned
only implicitly, without explicit world-consistency constraints. Consequently, static
structures may slide, warp, or flicker across viewpoints, while dynamic subjects
may exhibit implausible motion and appearance changes, as shown in
Fig.~\ref{fig:teaser}. Achieving world consistency therefore
remains a central challenge for faithful visual world simulation.

Existing efforts toward world-consistent video generation broadly follow two
routes. The first modifies the generator architecture to integrate explicit
geometric information, including point-conditioned generation
\citep{gen3c,viewcrafter,reconx,spatia} and joint RGB-geometry modeling
\citep{wvd,geovideo,fantasyworld,4dnex}. Although effective, the additional
modules and geometric supervision may compromise the broad generalization
inherited from large-scale video pretraining. The second preserves the
pretrained generator and improves consistency through post-training alignment,
allowing expert knowledge or world priors to provide feedback while retaining
its broad generative capability.
\revision{However, real-world scenes often contain both static environments and
dynamic subjects, making reliable consistency feedback in such scenes a key
challenge for post-training. Epipolar-DPO, VideoGPA, and the geometric rewards
of World-R1 rely on static-scene assumptions
\citep{epipolar-dpo,videogpa,worldr1}. In dynamic scenes, independently moving
subjects violate these assumptions, causing legitimate subject motion to be
penalized and potentially driving the generator toward frozen subjects.
VGGRPO and GeoFlow accommodate dynamic scenes by using motion cues to exclude
or downweight dynamic regions during static-consistency evaluation
\citep{vggrpo,geoflow}. However, alongside legitimate subject motion, these
cues also identify unintended sliding or warping of static regions as dynamic,
weakening the static-consistency signal (see Fig.~\ref{fig:motion_region_comparison}). Dynamic consistency is also
insufficiently addressed: VGGRPO lacks dedicated feedback for dynamic subjects,
while GeoFlow focuses on appearance consistency without explicitly assessing
whether subjects move as intended or whether their motion is physically
plausible. These limitations motivate us to develop more reliable and
comprehensive world-consistency rewards.}

\begin{figure}[!t]
    \centering
    \includegraphics[width=\textwidth]{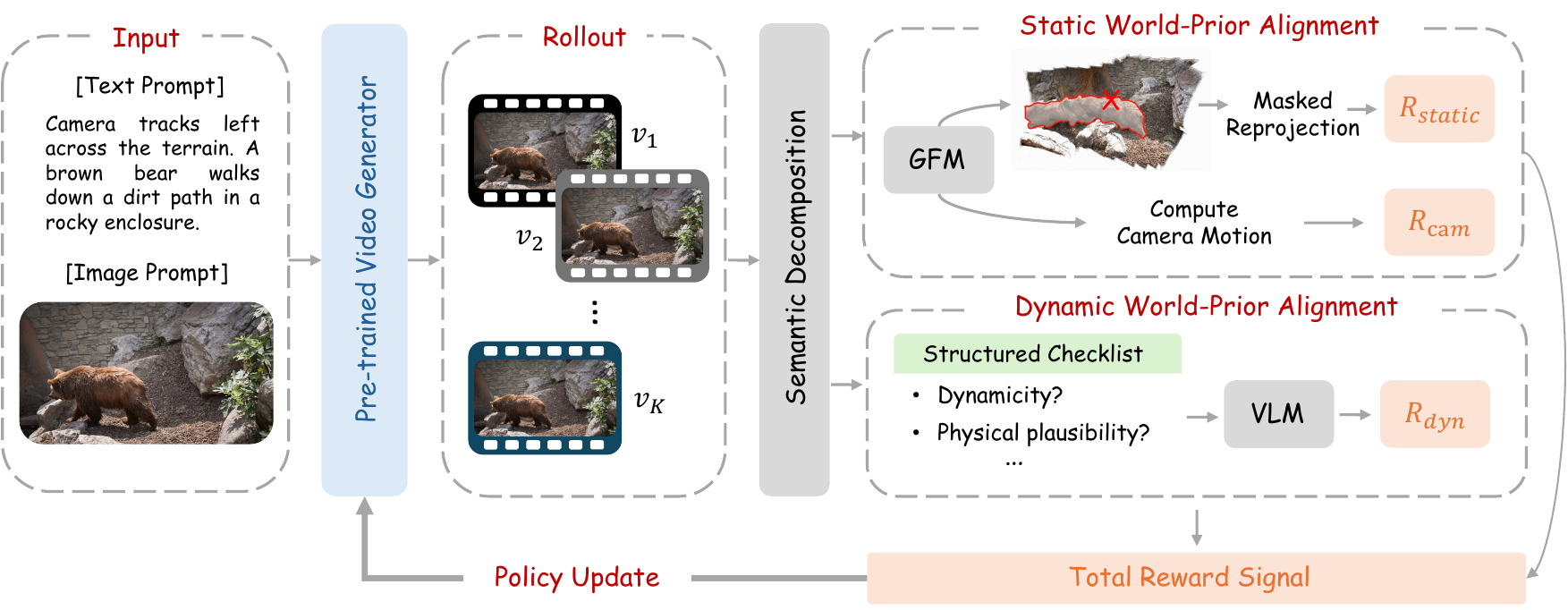}
    \caption{Overview of the proposed WorldAlign. Given generated
    videos, WorldAlign semantically segments and tracks dynamic subjects to
    decouple each video into static and dynamic regions. Static regions are
    aligned with a geometric world prior through masked reprojection and
    camera-motion rewards, while dynamic subjects are aligned with a VLM-based
    world prior through a structured \emph{VLM-as-a-judge} reward. The reward signals are normalized
    within each same-prompt sample group and aggregated for online policy
    update.}
    \label{fig:method_framework}
    \vspace{-5mm}
\end{figure}

To this end, we introduce \emph{WorldAlign} (Fig.~\ref{fig:method_framework}), a decoupled
4D reward framework that semantically separates static regions and dynamic subjects of each generated video and
provides feedback by aligning each with a world prior suited to its assumptions, rather
than imposing a static-scene assumption on the entire scene. For static consistency,
WorldAlign first semantically segments and tracks dynamic subjects rather than relying on
motion cues. This enables geometric evaluation to exclude independently moving subjects
while retaining static regions, including those exhibiting unintended sliding or warping.
WorldAlign then uses a geometry foundation model~\citep{vggt} as a geometric world prior
to construct a shared point cloud given the generated video and reproject it into the input views. Dynamic subjects
are masked out during both point-cloud construction and reprojection-error computation,
reducing their interference with static-consistency evaluation. When the frames depict a
coherent static 3D scene, the shared geometry should reproduce their static regions
consistently, whereas inconsistent structures lead to larger reprojection errors. The
resulting masked reprojection reward therefore penalizes inconsistent static structures
without penalizing legitimate subject motion. Since reducing camera motion can make
geometric consistency easier to satisfy, an auxiliary camera-motion reward discourages
nearly static solutions and encourages meaningful viewpoint changes. For dynamic consistency, WorldAlign explicitly evaluates their
motion and appearance over time. Because plausible subject evolution depends on the
subject's properties and scene context, we use a strong vision-language model (VLM) as a
dynamic world prior. We construct a VLM-as-a-judge reward based on sample-specific
checklists that translate this world knowledge into explicit criteria covering dynamicity,
physical plausibility, shape consistency, and texture consistency. The fraction of
satisfied checklist items provides dynamic-consistency feedback for each video that evaluates not only
appearance preservation but also whether subjects move as intended and whether their
motion is physically plausible. Together, these rewards enable online post-training to
jointly improve static and dynamic consistency without modifying the generator
architecture or requiring human preference annotations.

\revision{To validate WorldAlign, we compare it with state-of-the-art post-training
methods across two pretrained image-to-video generators using both reward-related
and reward-independent metrics. Experiments demonstrate that WorldAlign improves
static geometry and dynamic consistency over existing methods without
suppressing camera or subject motion. These results support decoupled world-prior
alignment as an effective approach to more faithful visual world simulation.}

\section{Related Work}
\subsection{Pretrained Video Generation Models}
Large-scale pretrained video generation models
\citep{sora,seedance2,klingomni,hunyuanvideo,wan} have substantially advanced
open-domain video synthesis. Their progress is driven by diffusion \citep{ddpm}
and flow-matching objectives \citep{flow-matching,rectified-flow},
transformer-based architectures \citep{vit}, and large-scale video-text training.
However, their pretraining objectives primarily optimize prediction in RGB or
latent space, leaving 3D structure and physical dynamics to be learned only
implicitly, without explicit world-consistency constraints. Consequently, world
consistency remains unreliable across generations. WorldAlign therefore explicitly
optimizes the world consistency of these pretrained models through post-training.

\subsection{World-Consistent Video Generation}
Existing approaches improve world consistency through two main routes:
\ding{172} \textbf{architecture modification for geometry integration},
which modifies the generator architecture to incorporate geometric information
into the generation process, and
\ding{173} \textbf{geometry-aware post-training},
which derives geometric feedback from generated videos to refine pretrained
generators through post-training.

\paragraph{Architecture Modification for Geometry Integration.}
Methods in this route introduce additional conditioning inputs derived from
reconstructed 3D points and train the generator to use them
\citep{gen3c,viewcrafter,reconx,spatia}. Other methods extend the generator to
jointly predict RGB appearance and geometric representations such as depth or
3D coordinates \citep{wvd,geovideo,fantasyworld,4dnex}. These designs require
architectural modifications and additional training to accommodate geometric
inputs or outputs, which may compromise the broad generalization acquired
through large-scale pretraining. WorldAlign instead improves the world consistency through post-training alignment, leaving the generator
architecture unchanged.

\paragraph{Geometry-Aware Post-Training.}
Epipolar-DPO and VideoGPA construct preference pairs from generated videos for
post-training, with preferences determined by epipolar error in
Epipolar-DPO~\citep{epipolar-dpo} and reprojection errors computed from predicted geometry
and camera poses in VideoGPA~\citep{videogpa}. Both geometric criteria rely on
static-scene assumptions; applying them directly to dynamic scenes can penalize legitimate
subject motion and encourage the generator to produce stationary subjects.
World-R1~\citep{worldr1} instead guides post-training with a 3D-aware reward alongside a
general visual reward, but this 3D-aware reward still relies on static-scene assumptions.
To accommodate dynamic content, it periodically disables the 3D-aware reward and optimizes
dynamic scenes using only the general visual reward, which provides no explicit
world-consistency constraints. VGGRPO and GeoFlow can provide geometric feedback for
dynamic scenes. They use motion cues to identify dynamic regions and reduce interference
from them during static-consistency evaluation. Specifically,
VGGRPO~\citep{vggrpo} uses scene flow to filter out dynamic regions, while
GeoFlow~\citep{geoflow} uses optical-flow residuals to downweight their contributions.
However, these cues can also identify unintended sliding or warping of static regions as
dynamic, weakening the static-consistency signal by excluding or downweighting regions
whose errors should be penalized. Dynamic consistency also remains insufficiently
addressed: VGGRPO provides no dedicated consistency feedback for the filtered subjects,
while GeoFlow measures DINO-based appearance consistency across the whole scene after flow
alignment. This appearance-based evaluation does not explicitly assess whether subjects
move as intended or whether their motion is physically plausible. To improve world
consistency in real-world dynamic scenes, WorldAlign uses tracked semantic masks to
exclude independently moving subjects from static-consistency evaluation while retaining
inconsistent static regions, providing more reliable static-consistency feedback than
methods based on motion cues. It also provides dedicated feedback for dynamic subjects
through VLM-as-a-judge reward that assesses dynamicity, physical plausibility, shape
consistency, and texture consistency.

\section{Method}
\subsection{Method Overview}
We consider image-to-video generation\footnote{WorldAlign can be extended to
text-to-video generation by obtaining subject masks from generated videos and
constructing checklists from text prompts.}, where a pretrained generator
produces an $N$-frame video $v=\{I_n\}_{n=1}^{N}$ conditioned on an image $I^c$
and a text prompt $y$. Our goal is to improve the world consistency of generated
videos through post-training without modifying the generator architecture.

Real-world scenes contain both static environments and dynamic subjects, which
satisfy different consistency assumptions. WorldAlign therefore separates them and provides feedback by aligning each with a suitable world prior, as illustrated in Fig.~\ref{fig:method_framework}. Given a generated video $v$,
we initialize a segmentation tracker~\cite{sam3} with the subject mask $M^c$ of
the conditioning image. The tracker propagates the mask across the video to
obtain subject masks $M=\{M_n\}_{n=1}^{N}$, which define the dynamic regions,
while their complements $\overline{M}_n$ define the static regions. For static
regions, a geometry foundation model is used as the prior for a masked
reprojection reward $R_{\mathrm{static}}$, while an auxiliary camera-motion
reward $R_{\mathrm{cam}}$ discourages nearly static solutions. For dynamic
subjects, a vision-language model is used as the prior for a structured
\emph{VLM-as-a-judge} reward $R_{\mathrm{dyn}}$. These rewards jointly
guide online post-training through GRPO to update the generator.

\subsection{Static World-Prior Alignment}
\leavevmode

\begin{wrapfigure}{r}{0.48\textwidth}
    \centering
    \includegraphics[width=\linewidth]{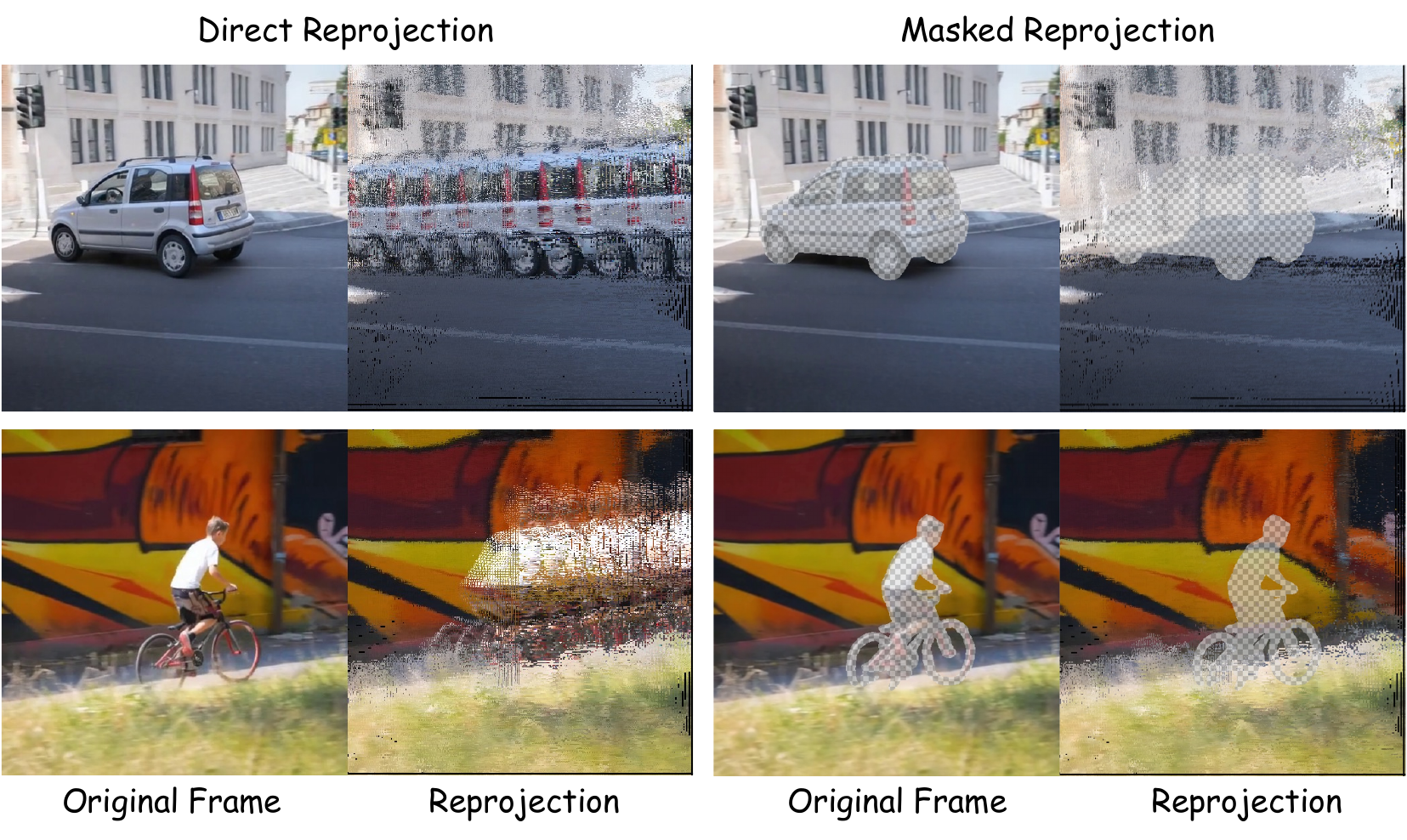}
    \caption{Motion-induced artifacts of Direct Reprojection.
    Our masked reprojection can isolate static geometry from subject
    motion, yielding a more reliable static-consistency reward.}
    
    \label{fig:motion_artifacts}
\end{wrapfigure}
\textbf{Masked Reprojection Reward.}
Static regions approximately satisfy rigid multi-view geometry. We therefore
use a geometry foundation model (GFM)~\citep{vggt} to measure their consistency
through cross-view reprojection. Given a generated video $v$, the GFM predicts a
pixel-aligned 3D point map $P_n$, camera intrinsics $K_n$, and camera extrinsics
$C_n$ for each frame $I_n$.

However, directly aggregating the predicted point maps also fuses the dynamic
subject at its successive positions into the shared point cloud, forming a 3D
trace of its motion. Directly reprojecting this point cloud into each frame produces
motion-induced artifacts. Pixels affected by these artifacts yield large
reprojection errors regardless of whether the corresponding static regions are
geometrically consistent, making direct reprojection unreliable for
static-consistency evaluation, as shown in
Fig.~\ref{fig:motion_artifacts}.

To remove this interference, WorldAlign constructs the shared point cloud using
only static regions. Because $P_n$ is pixel-aligned, the tracked mask $M_n$ can
be applied directly to exclude 3D points corresponding to the dynamic subject:
\begin{equation}
\mathcal{P}_{\mathrm{static}} =
\left\{
P_n(x)
\mid x\in\overline{M}_n,\; n=1,\ldots,N
\right\}.
\end{equation}
The resulting point cloud excludes the subject's motion trajectory and retains
only the 3D points assigned to static regions.
\revision{Because dynamic regions are defined by tracked subject masks rather than
motion cues, undesired distortions in static regions still remain during static-consistency evaluation.}
WorldAlign then projects $\mathcal{P}_{\mathrm{static}}$ into each frame using
the predicted camera parameters:
$\hat{I}_n = \Pi\left(\mathcal{P}_{\mathrm{static}};K_n,C_n\right)$,
where $\Pi$ denotes the reprojection operation. Although the reprojected frames $\hat{I}_n$
are rendered from the point cloud containing only the static environments, the original frames $I_n$ still contain
dynamic subjects. We therefore apply the same subject mask to each original
frame and its reprojected counterpart, excluding the corresponding pixel
locations from comparison. We further restrict the comparison with
valid reprojections. Thus, the valid comparison set is defined as
\begin{equation}
\Omega =
\left\{
(n,x)\mid x\in\overline{M}_n,\; \hat{I}_n(x)\ \text{is valid}
\right\}.
\end{equation}
Here, $n$ indexes frames and $x$ denotes a pixel location. A pair $(n,x)$ belongs
to $\Omega$ only if $x$ lies outside the subject mask in frame $n$ and has a valid
value in the corresponding reprojected frame.
The masked reprojection error and its corresponding static-consistency reward are
\begin{equation}
E_{\mathrm{rep}} =
\frac{1}{|\Omega|}
\sum_{(n,x)\in\Omega}
\left\|I_n(x)-\hat{I}_n(x)\right\|_2^2,
\enspace
R_{\mathrm{static}}=-E_{\mathrm{rep}}.
\end{equation}

\noindent\textbf{Camera-Motion Reward.}
The masked reprojection reward admits a trivial solution: suppressing viewpoint
changes can reduce the reprojection error without requiring consistent 3D
structure. To prevent this collapse, WorldAlign reuses the camera extrinsics
predicted by the GFM to construct an auxiliary reward that penalizes
insufficient camera motion.

Let $C_n=(R_n,t_n)$ denote the predicted camera extrinsics, with rotation
$R_n$ and translation $t_n$. We define the camera-motion magnitude as
$m(C_{1:N})=\bar{d}_{\mathrm{trans}}+\bar{d}_{\mathrm{rot}}$,
where $\bar{d}_{\mathrm{trans}}$ and $\bar{d}_{\mathrm{rot}}$ denote the mean
adjacent-frame translation $\|t_{n+1}-t_n\|_2$ and mean adjacent-frame rotation angle
$\theta(R_{n+1}R_n^\top)$, respectively.
The camera-motion reward applies a hinge penalty when the predicted motion falls
below a threshold $\tau$:
\begin{equation}
R_{\mathrm{cam}}
=
-\max\left(0,\tau-m(C_{1:N})\right),
\end{equation}
where $\tau$ is the lower-bound camera-motion threshold. This reward penalizes
videos with insufficient viewpoint changes, preventing the generator from
reducing the reprojection error solely through nearly static viewpoints.

\subsection{Dynamic World-Prior Alignment}
\revision{WorldAlign uses a strong VLM as a dynamic world prior and translates
its world knowledge into the \emph{VLM-as-a-judge} reward based on subject-specific binary questions.}

Before training, we use a VLM to construct a sample-specific checklist
$\mathcal{Q}=\{q_j\}_{j=1}^{J}$ from each conditioning pair $(I^c,y)$, where $J$ is the number of questions in each checklist. The
checklist consists of subject-centric binary questions tailored to the subject
and its intended motion, covering four complementary aspects of dynamic
consistency:
\begin{itemize}
    \item \textbf{Dynamicity}: whether the subject exhibits the intended motion
    rather than remaining static.
    \item \textbf{Physical plausibility}: whether the motion follows physically
    plausible dynamics given the subject's intrinsic properties and the
    surrounding environment.
    \item \textbf{Shape}: whether the subject preserves its structure over time
    while allowing physically plausible articulation.
    \item \textbf{Texture}: whether the subject preserves its defining visual
    attributes, such as color, patterns, and surface details.
\end{itemize}
Here, the shape and texture criteria prevent appearance drift. The sample-specific checklist
accounts for subject- and context-dependent dynamics, while its binary format
decomposes an open-ended judgment into explicit criteria for more reliable
evaluation.

During training, the judging VLM receives the full generated video $v$ and the
checklist $\mathcal{Q}$. The full video allows it to consider the surrounding
scene, while the evaluation instruction directs it to focus on the dynamic
subject and answer each question with YES or NO. Let $a_j\in\{0,1\}$ denote the
answer to question $q_j$, with YES mapped to $1$. We define the
\emph{VLM-as-a-judge} reward as
\begin{equation}
R_{\mathrm{dyn}}=
\frac{1}{J}\sum_{j=1}^{J}a_j.
\end{equation}
Thus, $R_{\mathrm{dyn}}$ measures the fraction of checklist criteria satisfied
by the generated video.

\subsection{Online Policy Update}
We optimize the pretrained generator using online GRPO following
Flow-GRPO~\citep{flow-grpo}. For each conditioning pair $(I^c,y)$, the generator
samples a group of $K$ videos $\{v_i\}_{i=1}^{K}$. For every $v_i$, WorldAlign
computes the three reward components $R_{\mathrm{static}}$,
$R_{\mathrm{cam}}$, and $R_{\mathrm{dyn}}$.

To obtain group-relative signals while balancing their different scales, we
normalize each reward component separately within the group:
\begin{equation}
\widehat{R}_{k}(v_i)=
\frac{R_k(v_i)-\mu_k}{\sigma_k+\epsilon},
\qquad
k\in\{\mathrm{static},\mathrm{cam},\mathrm{dyn}\},
\end{equation}
where $\mu_k$ and $\sigma_k$ are the group mean and standard deviation of reward
$k$, and $\epsilon$ is a small constant for numerical stability. The normalized
components are aggregated as the advantage:
\begin{equation}
A_i =
\lambda_s\widehat{R}_{\mathrm{static}}(v_i)
+\lambda_c\widehat{R}_{\mathrm{cam}}(v_i)
+\lambda_d\widehat{R}_{\mathrm{dyn}}(v_i),
\end{equation}
where $\lambda_s$, $\lambda_c$, and $\lambda_d$ control the contributions of
the three reward components.
The resulting advantage is used with the standard clipped GRPO objective to
update the generator. The tracker, GFM, and VLM remain frozen and are used only
for reward computation.

\section{Experiments}

\subsection{Experimental Setup}
\noindent\textbf{Datasets.}
WorldAlign focuses on real-world dynamic scenes in which static environments and dynamic
subjects coexist. Existing datasets, such as DL3DV~\citep{dl3dv}, contain
only static scenes and therefore cannot support dynamic consistency
alignment. We thus construct training and evaluation dataset from
DAVIS~\citep{davis} and COCO~\citep{mscoco} for real-world dynamic scenes. Each sample comprises a conditioning
image, a subject mask for the conditioning image, an image-to-video prompt, and
a sample-specific dynamic-consistency checklist.
The resulting dataset contains 1,389 samples, of which 100 are randomly held out
for evaluation and the remaining 1,289 are used for training. Further details on
data selection, prompt generation, and checklist construction are provided in
\hyperref[sec:arxiv_supp_dataset]{Appendix~\ref*{sec:arxiv_supp_dataset}}.

\noindent\textbf{Baselines.}
We compare WorldAlign with the original pretrained base model and four post-training
methods: Epipolar-DPO~\citep{epipolar-dpo}, VideoGPA~\citep{videogpa},
World-R1~\citep{worldr1}, and GeoFlow~\citep{geoflow}. These comparisons are conducted
separately on two base models, Wan2.2-TI2V-5B and Wan2.1-I2V-14B-480P~\citep{wan}. For
each base model, all post-training methods use the same training data and training
settings for a fair comparison. We reproduce World-R1 for our setting without explicit
camera trajectories. Our reproduction retains its reconstruction, meta-view, and general
visual rewards, but omits camera-aware noise warping and trajectory-alignment rewards.
VGGRPO~\citep{vggrpo} is excluded because its official implementation is not publicly
available.

\noindent\textbf{Evaluation Metrics.}
We evaluate generated videos along three dimensions: static consistency,
dynamic consistency, and general video quality. For static and dynamic
consistency, we pair reward-related and reward-independent metrics to verify
gains beyond the optimized objective. General video
quality is assessed using third-party VBench metrics~\citep{vbench}.
\par\noindent\ding{172} \emph{Static Consistency.} We report {\textbf{Masked Epipolar Error (M.Epi.)}}
and {\textbf{Masked Reprojection Error (M.Reproj.)}}. M.Epi. is
reward-independent: it measures epipolar violations between static-region
correspondences. M.Reproj. measures the discrepancy between original and
reprojected static regions.
Both exclude tracked dynamic subjects. Lower values are better. Because
suppressing viewpoint changes can trivially reduce both errors, we interpret
them jointly with Dynamic Degree.
\par\noindent\ding{173} \emph{Dynamic Consistency.} We report {\textbf{Tracked-Subject DINOv3
Similarity (Subject DINO)}} and the {\textbf{Checklist ``Yes'' Rate}}. Tracked-Subject DINOv3
Similarity~\citep{dinov3} is reward-independent and computes the mean
cosine similarity of DINOv3 features in tracked subject regions across adjacent
frames. Checklist ``Yes'' Rate measures the proportion of dynamic-consistency
criteria satisfied by each video. Higher values are better.
Because a frozen subject can trivially produce high DINO similarity, we
interpret it jointly with Dynamicity ``Yes'' Rate and Dynamic Degree.
\par\noindent\ding{174} \emph{General Video Quality.} Following VBench~\citep{vbench}, we
report \textbf{Dynamic Degree}, \textbf{Motion Smoothness}, and \textbf{Aesthetic Quality}, with higher
values being better. Dynamic Degree measures the overall amount of motion and
helps identify motion suppression, while Motion Smoothness and Aesthetic
Quality assess motion continuity and frame-wise visual quality, respectively.

\textcolor{black}{\textbf{Implementation details} are provided in \hyperref[sec:arxiv_supp_implementation]{Appendix~\ref*{sec:arxiv_supp_implementation}}.}

\subsection{Comparison with State-of-the-Art Methods}
\textcolor{black}{We compare WorldAlign with pretrained models and geometry-aware alignment methods on static consistency, dynamic consistency, and motion preservation.}
Complementary \textbf{human evaluation} is provided in
\hyperref[sec:arxiv_supp_human_evaluation]{Appendix~\ref*{sec:arxiv_supp_human_evaluation}}
to further support the quantitative results in Tab.~\ref{tab:sota_comparison}.

As shown in Tab.~\ref{tab:sota_comparison}, the pretrained base models retain
moderate motion but exhibit high geometric errors and limited dynamic
consistency, confirming that world consistency remains an unreliable capability
of pretrained video generators. 
Epipolar-DPO and VideoGPA report deceptively low geometric errors, accompanied
by substantial decreases in both Dynamic Degree and Dynamicity ``Yes'' Rate.
Their apparent geometric gains are therefore largely attributable to the
suppression of camera and subject motion rather than real improvements in world
consistency. VideoGPA exhibits less severe motion suppression because it filters
candidates with insufficient camera motion during preference construction,
whereas Epipolar-DPO has no such mechanism. Our World-R1 reproduction provides
only marginal improvements in static consistency while still suppressing
subject motion, as reflected by its lower Dynamic Degree and Dynamicity ``Yes''
Rate. Its reconstruction and meta-view terms evaluate the reconstructed scene
through perceptual and semantic similarity without isolating dynamic subjects.
In dynamic scenes, subject-motion discrepancies can therefore dominate the
reward, diluting the static-consistency signal and biasing optimization toward
subject-motion suppression. 
GeoFlow improves static and dynamic consistency over the base models but remains
less effective than WorldAlign. Its motion cues may downweight inconsistent static
regions, weakening static-consistency feedback, while its appearance-based reward
does not explicitly assess motion plausibility. WorldAlign addresses both
limitations and achieves stronger consistency without suppressing overall or
subject motion.
WorldAlign's gains in Motion
Smoothness and Aesthetic Quality further indicate the resulting dynamics
remain coherent and visually plausible.

\begin{table}[!t]
  \caption{Comparison with base models and state-of-the-art alignment
  methods on two image-to-video generators. For the Checklist
  ``Yes'' Rate, parentheses report the Dynamicity ``Yes'' Rate over all test
  videos. M.Reproj. and Checklist ``Yes'' Rate are reward-related; M.Epi. and
  Subject DINO are reward-independent. Bold and underlined entries denote the
  \textcolor{black}{best and second-best results, respectively.} Consistency metrics are interpreted
  jointly with Dynamic Degree because motion suppression can yield deceptively
  favorable scores, shown in \textcolor{gray}{gray}. WorldAlign improves both
  reward-related and reward-independent consistency metrics while preserving
  overall and subject motion.}
  \label{tab:sota_comparison}
  \centering
  {\footnotesize
  \setlength{\tabcolsep}{1pt}
  \renewcommand{\arraystretch}{1.05}
  \begin{tabularx}{\textwidth}{@{}>{\raggedright\arraybackslash}p{0.22\textwidth}*{3}{>{\hsize=.9\hsize\centering\arraybackslash}X}>{\hsize=1.6\hsize\centering\arraybackslash}X*{3}{>{\hsize=.9\hsize\centering\arraybackslash}X}@{}}
    \toprule
    & \multicolumn{2}{c}{Static Consistency}
    & \multicolumn{2}{c}{Dynamic Consistency}
    & \multicolumn{3}{c}{VBench} \\
    \cmidrule(lr){2-3}\cmidrule(lr){4-5}\cmidrule(l){6-8}
    Method &
    \shortstack{M.Epi.\\$\downarrow$} &
    \shortstack{M.Reproj.\\$\downarrow$} &
    \shortstack{Subject\\DINO $\uparrow$} &
    \shortstack{Checklist\\``Yes'' Rate $\uparrow$} &
    \shortstack{Dynamic\\Degree$\uparrow$} &
    \shortstack{Motion\\Smooth. $\uparrow$} &
    \shortstack{Aesthetic\\Quality $\uparrow$} \\
    \midrule
    \multicolumn{8}{@{}l}{\emph{Wan2.2-TI2V-5B}} \\
    \rowcolor{BaseRowGray}
    Base                         & 0.463 & 0.040 & 0.922 & \textcolor{black}{0.697 (0.63)} & \textcolor{black}{0.63} & 0.957 & 0.479 \\
    Epipolar-DPO & \textbf{\textcolor{gray}{{0.268}}} & \textbf{\textcolor{gray}{0.022}} & \textbf{\textcolor{gray}{0.961}} & 0.646 (0.43) & 0.31 & \underline{\textcolor{gray}{0.975}} & \underline{0.481} \\
    VideoGPA    & \underline{\textcolor{gray}{{0.308}}} & \underline{\textcolor{gray}{0.024}} & \underline{\textcolor{gray}{0.948}} & 0.652 (0.56) & 0.54 & \textcolor{gray}{0.971} & 0.476 \\
    World-R1     & 0.432 & 0.038 & \textcolor{gray}{0.945} & 0.657 (0.61) & 0.56 & 0.965 & 0.459 \\
    GeoFlow & \textcolor{black}{0.367} & \textcolor{black}{0.031} & \textcolor{black}{0.934} & \textcolor{black}{\underline{0.736 (0.74)}} & \textcolor{black}{\underline{0.67}} & \textcolor{black}{0.971} & \textcolor{black}{0.478} \\
    \rowcolor{OursRow}
    \textbf{WorldAlign (Ours)}   & 0.321 & 0.026 & 0.938 & \textbf{0.781 (0.85)} & \textbf{0.72} & \textbf{0.979} & \textbf{0.488} \\
    \midrule
    \multicolumn{8}{@{}l}{\emph{Wan2.1-I2V-14B-480P}} \\
    \rowcolor{BaseRowGray}
    Base                         & 0.469 & 0.037 & 0.918 & \textcolor{black}{0.760 (0.81)} & \textcolor{black}{0.62} & 0.963 & 0.481 \\
    Epipolar-DPO & \textbf{\textcolor{gray}{0.215}} & \textbf{\textcolor{gray}{0.019}} & \textbf{\textcolor{gray}{0.967}} & 0.594 (0.53) & 0.33 & \underline{\textcolor{gray}{0.976}} & \textcolor{black}{\textbf{0.494}} \\
    VideoGPA    & \underline{\textcolor{gray}{0.243}} & \underline{\textcolor{gray}{0.022}} & \underline{\textcolor{gray}{0.963}} & 0.608 (0.61) & 0.42 & \textcolor{gray}{0.970} & 0.489 \\
    World-R1     & 0.428 & 0.034 & \textcolor{gray}{0.951} & 0.636 (0.72) & 0.53 & 0.962 & 0.440 \\
    GeoFlow & \textcolor{black}{0.336} & \textcolor{black}{0.028} & \textcolor{black}{0.936} & \textcolor{black}{\underline{0.807 (0.83)}} & \textcolor{black}{\underline{0.65}} & \textcolor{black}{0.975} & \textcolor{black}{0.485} \\
    \rowcolor{OursRow}
    \textbf{WorldAlign (Ours)}   & 0.264 & 0.024 & 0.939 & \textbf{0.856 (0.92)} & \textbf{0.74} & \textbf{0.978} & \textcolor{black}{\underline{0.491}} \\
    \bottomrule
  \end{tabularx}
  \vspace{-5mm}
  }
\end{table}

\begin{table}[!t]
  \caption{Ablation of the proposed WorldAlign on Wan2.2-TI2V-5B.
  Camera and dynamic rewards are incrementally added to both the direct- and
  masked-reprojection branches. The results show that masked reprojection avoids
  imposing a static-scene assumption on the entire video and prevents the
  suppression of dynamic subject motion. The camera reward alleviates the static
  collapse observed with masked reprojection alone, while the dynamic reward
  further improves dynamic consistency.}
  \label{tab:reward_ablation}
  \centering
  {\footnotesize
  \setlength{\tabcolsep}{1pt}
  \renewcommand{\arraystretch}{1.05}
  \begin{tabularx}{\textwidth}{@{}>{\raggedright\arraybackslash}p{0.22\textwidth}*{3}{>{\hsize=.9\hsize\centering\arraybackslash}X}>{\hsize=1.6\hsize\centering\arraybackslash}X*{3}{>{\hsize=.9\hsize\centering\arraybackslash}X}@{}}
    \toprule
    & \multicolumn{2}{c}{Static Consistency}
    & \multicolumn{2}{c}{Dynamic Consistency}
    & \multicolumn{3}{c}{VBench} \\
    \cmidrule(lr){2-3}\cmidrule(lr){4-5}\cmidrule(l){6-8}
    Variant &
    \shortstack{M.Epi.\\$\downarrow$} &
    \shortstack{M.Reproj.\\$\downarrow$} &
    \shortstack{Subject\\DINO $\uparrow$} &
    \shortstack{Checklist\\``Yes'' Rate $\uparrow$} &
    \shortstack{Dynamic\\Degree $\uparrow$} &
    \shortstack{Motion\\Smooth. $\uparrow$} &
    \shortstack{Aesthetic\\Quality $\uparrow$} \\
    \midrule
    \rowcolor{BaseRowGray}
    Base & 0.463 & 0.0396 & 0.922 & 0.697 (0.63) & 0.63 & 0.957 & 0.479 \\
    \midrule
    w/ Direct R. Reward & \underline{\textcolor{gray}{0.298}} & \underline{\textcolor{gray}{0.0223}}
      & \textbf{\textcolor{gray}{0.968}} & 0.628 (0.18) & 0.24 & \textbf{\textcolor{gray}{0.992}}
      & \underline{0.482} \\
    + Camera Reward & 0.324 & 0.0272 & \underline{\textcolor{gray}{0.963}} & 0.632 (0.22)
      & 0.64 & \textcolor{gray}{0.982} & 0.479 \\
    + Dynamic Reward & 0.338 & 0.0291 & \textcolor{gray}{0.954} & 0.670 (0.55)
      & 0.69 & \textcolor{gray}{0.974} & 0.477 \\
    \midrule
    w/ Masked R. Reward & \textbf{\textcolor{gray}{0.289}} & \textbf{\textcolor{gray}{0.0215}} & \textcolor{gray}{0.956}
      & 0.667 (0.55) & 0.42 & \underline{\textcolor{gray}{0.986}} & 0.480 \\
    + Camera Reward & 0.317 & 0.0262 & 0.927
      & \underline{0.711 (0.64)} & \underline{0.70} & 0.968 & 0.475 \\
    \rowcolor{OursRow}
    \textbf{+ Dynamic Reward} & 0.321 & 0.0260
      & 0.938 & \textbf{0.781 (0.85)} & \textbf  {0.72}
      & 0.979 & \textbf{0.488} \\
    \bottomrule
  \end{tabularx}
  }
\end{table}

\subsection{Ablation Study}

\textcolor{black}{Tab.~\ref{tab:reward_ablation} evaluates the reward components.
Masked reprojection alone reduces geometric errors but suppresses motion;
adding the Camera Reward raises Dynamic Degree from 0.42 to 0.70.
The Dynamic Reward further improves Subject DINO and Checklist ``Yes'' Rate
while retaining the geometric gains. Direct reprojection instead penalizes
legitimate subject motion, creating conflicting objectives and yielding poorer
static and dynamic consistency than the full masked design.
The complete analysis is provided in \hyperref[sec:arxiv_supp_ablation]{Appendix~\ref*{sec:arxiv_supp_ablation}}.}

\subsection{\textcolor{black}{Comparison of Dynamic Region Identification}}

\begin{figure}[!t]
    \centering
    \vspace{-3mm}
    \includegraphics[width=\textwidth]{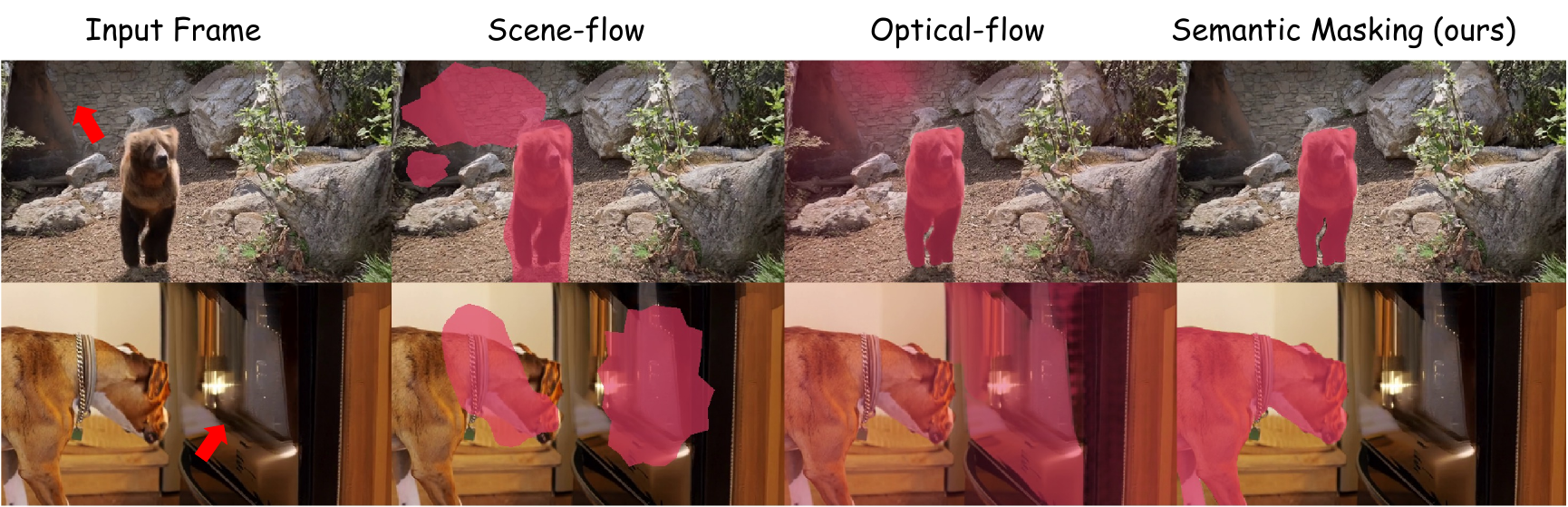}
    \caption{\textcolor{black}{Qualitative comparison of dynamic region identification.
    The first column shows only the first frame of each video; red arrows mark
    static regions that distort over time and should remain during static
    consistency evaluation. Red overlays indicate regions identified as dynamic.
    Scene flow and optical flow, the motion cues used by VGGRPO and GeoFlow,
    respectively, also mark distorted static regions,
    whereas our semantic masks more cleanly isolate the dynamics.}}
    \vspace{-3mm}
    \label{fig:motion_region_comparison}
\end{figure}

\textcolor{black}{Fig.~\ref{fig:motion_region_comparison} compares dynamic region
identification using scene flow and optical flow, the motion cues used by
VGGRPO~\citep{vggrpo} and GeoFlow~\citep{geoflow}, respectively, with our semantic masking
on the same videos. In the first column, red arrows mark static regions that undergo
unintended distortions over time. These regions should remain included in
static-consistency evaluation, while dynamic subjects, such as the bear and dog, should be
excluded to avoid interference. Red overlays in the remaining three columns indicate
regions identified as dynamic by each method. Both motion cues also classify the distorted
static regions as dynamic. Consequently, filtering out or downweighting these regions
weakens the static-consistency signal for the errors that should be corrected. Our
semantic masks instead isolate and exclude dynamic subjects while retaining inconsistent
static regions for evaluation, providing more reliable static-consistency feedback.}

\section{Conclusion}
We introduced WorldAlign, a decoupled 4D reward framework that separates static
regions and dynamic subjects and aligns each with a world prior suited to its
assumptions. For static regions, a masked reprojection reward leverages a
geometry foundation model to assess static consistency without interference from
subject motion, while an auxiliary camera-motion reward prevents nearly static
solutions. For dynamic subjects, a \emph{VLM-as-a-judge} reward uses
sample-specific checklists to assess physical motion and appearance consistency.
These rewards enable online post-training without architectural changes or human
preference annotations. Across two pretrained image-to-video generators,
WorldAlign jointly improves static and dynamic consistency on both
reward-related and reward-independent metrics without suppressing overall or
subject motion. These results demonstrate the effectiveness of decoupled
world-prior alignment for advancing video generation toward more faithful visual
world simulation.

\section*{AI Use Statement}
Generative AI tools were used for manuscript language
polishing. The authors verified the cited references, reviewed all AI-assisted
edits, and take full responsibility for the final manuscript.

\bibliographystyle{iclr2027_conference}
\bibliography{references}

@article{qwen3,
  title={Qwen3 technical report},
  author={Yang, An and Li, Anfeng and Yang, Baosong and Zhang, Beichen and Hui, Binyuan and Zheng, Bo and Yu, Bowen and Gao, Chang and Huang, Chengen and Lv, Chenxu and others},
  journal={arXiv preprint arXiv:2505.09388},
  year={2025}
}

@inproceedings{vggt,
  title={Vggt: Visual geometry grounded transformer},
  author={Wang, Jianyuan and Chen, Minghao and Karaev, Nikita and Vedaldi, Andrea and Rupprecht, Christian and Novotny, David},
  booktitle={Proceedings of the Computer Vision and Pattern Recognition Conference},
  pages={5294--5306},
  year={2025}
}

@article{sam3,
  title={Sam 3: Segment anything with concepts},
  author={Carion, Nicolas and Gustafson, Laura and Hu, Yuan-Ting and Debnath, Shoubhik and Hu, Ronghang and Suris, Didac and Ryali, Chaitanya and Alwala, Kalyan Vasudev and Khedr, Haitham and Huang, Andrew and others},
  journal={arXiv preprint arXiv:2511.16719},
  year={2025}
}

@article{dinov3,
  title={Dinov3},
  author={Sim{\'e}oni, Oriane and Vo, Huy V and Seitzer, Maximilian and Baldassarre, Federico and Oquab, Maxime and Jose, Cijo and Khalidov, Vasil and Szafraniec, Marc and Yi, Seungeun and Ramamonjisoa, Micha{\"e}l and others},
  journal={arXiv preprint arXiv:2508.10104},
  year={2025}
}

@InProceedings{vbench,
      title={{VBench}: Comprehensive Benchmark Suite for Video Generative Models},
      author={Huang, Ziqi and He, Yinan and Yu, Jiashuo and Zhang, Fan and Si, Chenyang and Jiang, Yuming and Zhang, Yuanhan and Wu, Tianxing and Jin, Qingyang and Chanpaisit, Nattapol and Wang, Yaohui and Chen, Xinyuan and Wang, Limin and Lin, Dahua and Qiao, Yu and Liu, Ziwei},
      booktitle={Proceedings of the IEEE/CVF Conference on Computer Vision and Pattern Recognition},
      year={2024}
}

@inproceedings{dl3dv,
  title={Dl3dv-10k: A large-scale scene dataset for deep learning-based 3d vision},
  author={Ling, Lu and Sheng, Yichen and Tu, Zhi and Zhao, Wentian and Xin, Cheng and Wan, Kun and Yu, Lantao and Guo, Qianyu and Yu, Zixun and Lu, Yawen and others},
  booktitle={Proceedings of the IEEE/CVF Conference on Computer Vision and Pattern Recognition},
  pages={22160--22169},
  year={2024}
}

@article{gemini,
  title={Gemini: a family of highly capable multimodal models},
  author={Team, Gemini and Anil, Rohan and Borgeaud, Sebastian and Alayrac, Jean-Baptiste and Yu, Jiahui and Soricut, Radu and Schalkwyk, Johan and Dai, Andrew M and Hauth, Anja and Millican, Katie and others},
  journal={arXiv preprint arXiv:2312.11805},
  year={2023}
}

@inproceedings{mscoco,
  title={Microsoft coco: Common objects in context},
  author={Lin, Tsung-Yi and Maire, Michael and Belongie, Serge and Hays, James and Perona, Pietro and Ramanan, Deva and Doll{\'a}r, Piotr and Zitnick, C Lawrence},
  booktitle={European conference on computer vision},
  pages={740--755},
  year={2014},
  organization={Springer}
}

@article{davis,
  title={The 2017 davis challenge on video object segmentation},
  author={Pont-Tuset, Jordi and Perazzi, Federico and Caelles, Sergi and Arbel{\'a}ez, Pablo and Sorkine-Hornung, Alex and Van Gool, Luc},
  journal={arXiv preprint arXiv:1704.00675},
  year={2017}
}

@article{flow-grpo,
  title={Flow-grpo: Training flow matching models via online rl},
  author={Liu, Jie and Liu, Gongye and Liang, Jiajun and Li, Yangguang and Liu, Jiaheng and Wang, Xintao and Wan, Pengfei and Zhang, Di and Ouyang, Wanli},
  journal={Advances in neural information processing systems},
  volume={38},
  pages={40783--40818},
  year={2026}
}

@article{vit,
  title={An image is worth 16x16 words: Transformers for image recognition at scale},
  author={Dosovitskiy, Alexey and Beyer, Lucas and Kolesnikov, Alexander and Weissenborn, Dirk and Zhai, Xiaohua and Unterthiner, Thomas and Dehghani, Mostafa and Minderer, Matthias and Heigold, Georg and Gelly, Sylvain and others},
  journal={arXiv preprint arXiv:2010.11929},
  year={2020}
}

@article{rectified-flow,
  title={Flow straight and fast: Learning to generate and transfer data with rectified flow},
  author={Liu, Xingchao and Gong, Chengyue and Liu, Qiang},
  journal={arXiv preprint arXiv:2209.03003},
  year={2022}
}

@article{flow-matching,
  title={Flow matching for generative modeling},
  author={Lipman, Yaron and Chen, Ricky TQ and Ben-Hamu, Heli and Nickel, Maximilian and Le, Matt},
  journal={arXiv preprint arXiv:2210.02747},
  year={2022}
}

@article{ddpm,
  title={Denoising diffusion probabilistic models},
  author={Ho, Jonathan and Jain, Ajay and Abbeel, Pieter},
  journal={Advances in neural information processing systems},
  volume={33},
  pages={6840--6851},
  year={2020}
}

@article{videogpa,
  title={VideoGPA: Distilling Geometry Priors for 3D-Consistent Video Generation},
  author={Du, Hongyang and Ye, Junjie and Cong, Xiaoyan and Li, Runhao and Ni, Jingcheng and Agarwal, Aman and Zhou, Zeqi and Li, Zekun and Balestriero, Randall and Wang, Yue},
  journal={arXiv preprint arXiv:2601.23286},
  year={2026}
}

@article{epipolar-dpo,
  title={Epipolar Geometry Improves Video Generation Models},
  author={Kupyn, Orest and Manhardt, Fabian and Tombari, Federico and Rupprecht, Christian},
  journal={arXiv preprint arXiv:2510.21615},
  year={2025}
}

@article{vggrpo,
  title={Vggrpo: Towards world-consistent video generation with 4d latent reward},
  author={An, Zhaochong and Kupyn, Orest and Uscidda, Th{\'e}o and Colaco, Andrea and Ahuja, Karan and Belongie, Serge and Gonzalez-Franco, Mar and Gazulla, Marta Tintore},
  journal={arXiv preprint arXiv:2603.26599},
  year={2026}
}

@article{worldr1,
  title={World-r1: Reinforcing 3d constraints for text-to-video generation},
  author={Wang, Weijie and He, Xiaoxuan and Gu, Youping and Yang, Yifan and Zhang, Zeyu and He, Yefei and Ding, Yanbo and Hu, Xirui and Chen, Donny Y and He, Zhiyuan and others},
  journal={arXiv preprint arXiv:2604.24764},
  year={2026}
}

@article{dvgwm,
  title={DVG-WM: Disentangled Video Generation Enables Efficient Embodied World Model for Robotic Manipulation},
  author={Shan, Ziyu and Wu, Zhenyu and Wang, Xiaofeng and Zhu, Zheng and Wang, Ziwei},
  journal={arXiv preprint arXiv:2606.32028},
  year={2026}
}

@article{physisforcing,
  title={PhysisForcing: Physics Reinforced World Simulator for Robotic Manipulation},
  author={Zhang, Peiwen and Deng, Yufan and Sun, Shangkun and Ma, Juncheng and Wang, Duomin and Du, Jonas and Pan, Zilin and Huang, Ye and Liang, Hao and Huang, Songyan and others},
  journal={arXiv preprint arXiv:2606.28128},
  year={2026}
}

@article{gaia1,
  title={Gaia-1: A generative world model for autonomous driving},
  author={Hu, Anthony and Russell, Lloyd and Yeo, Hudson and Murez, Zak and Fedoseev, George and Kendall, Alex and Shotton, Jamie and Corrado, Gianluca},
  journal={arXiv preprint arXiv:2309.17080},
  year={2023}
}

@inproceedings{drivedreamer,
  title={Drivedreamer: Towards real-world-drive world models for autonomous driving},
  author={Wang, Xiaofeng and Zhu, Zheng and Huang, Guan and Chen, Xinze and Zhu, Jiagang and Lu, Jiwen},
  booktitle={European conference on computer vision},
  pages={55--72},
  year={2024},
  organization={Springer}
}

@article{vista,
  title={Vista: A generalizable driving world model with high fidelity and versatile controllability},
  author={Gao, Shenyuan and Yang, Jiazhi and Chen, Li and Chitta, Kashyap and Qiu, Yihang and Geiger, Andreas and Zhang, Jun and Li, Hongyang},
  journal={Advances in Neural Information Processing Systems},
  volume={37},
  pages={91560--91596},
  year={2024}
}

@inproceedings{genie,
  title={Genie: Generative interactive environments},
  author={Bruce, Jake and Dennis, Michael D and Edwards, Ashley and Parker-Holder, Jack and Shi, Yuge and Hughes, Edward and Lai, Matthew and Mavalankar, Aditi and Steigerwald, Richie and Apps, Chris and others},
  booktitle={Forty-first International Conference on Machine Learning},
  year={2024}
}

@article{dreamxworld,
title={DreamX-World 1.0: A General-Purpose Interactive World Model},
  author={Team, DreamX and Bai, Yancheng and Chen, Rui and Chu, Xiangxiang and Dang, Rujing and Dou, Hao and Gao, Bingjie and Gu, Qiwen and Hong, Siyu and Lei, Jiachen and others},
  journal={arXiv preprint arXiv:2606.16993},
  year={2026}
}

@article{hyworldplay,
title={Worldplay: Towards long-term geometric consistency for real-time interactive world modeling},
  author={Sun, Wenqiang and Zhang, Haiyu and Wang, Haoyuan and Wu, Junta and Wang, Zehan and Wang, Zhenwei and Wang, Yunhong and Zhang, Jun and Wang, Tengfei and Guo, Chunchao},
  journal={arXiv preprint arXiv:2512.14614},
  year={2025}
}

@article{sora,
  title={Video generation models as world simulators},
  author={Brooks, Tim and Peebles, Bill and Holmes, Connor and DePue, Will and Guo, Yufei and Jing, Leo and Schnurr, David and Taylor, Joe and Luhman, Troy and Luhman, Eric and others},
  journal={OpenAI Blog},
  volume={1},
  number={8},
  pages={1},
  year={2024}
}

@article{seedance2,
  title={Seedance 2.0: Advancing video generation for world complexity},
  author={Seedance, Team and Chen, De and Chen, Liyang and Chen, Xin and Chen, Ying and Chen, Zhuo and Chen, Zhuowei and Cheng, Feng and Cheng, Tianheng and Cheng, Yufeng and others},
  journal={arXiv preprint arXiv:2604.14148},
  year={2026}
}

@article{klingomni,
 title={Kling-Omni Technical Report},
  author={Team, Kling and Chen, Jialu and Ci, Yuanzheng and Du, Xiangyu and Feng, Zipeng and Gai, Kun and Guo, Sainan and Han, Feng and He, Jingbin and He, Kang and others},
  journal={arXiv preprint arXiv:2512.16776},
  year={2025}
}

@article{hunyuanvideo,
  title={Hunyuanvideo: A systematic framework for large video generative models},
  author={Kong, Weijie and Tian, Qi and Zhang, Zijian and Min, Rox and Dai, Zuozhuo and Zhou, Jin and Xiong, Jiangfeng and Li, Xin and Wu, Bo and Zhang, Jianwei and others},
  journal={arXiv preprint arXiv:2412.03603},
  year={2024}
}

@article{wan,
  title={Wan: Open and advanced large-scale video generative models},
  author={Wan, Team and Wang, Ang and Ai, Baole and Wen, Bin and Mao, Chaojie and Xie, Chen-Wei and Chen, Di and Yu, Feiwu and Zhao, Haiming and Yang, Jianxiao and others},
  journal={arXiv preprint arXiv:2503.20314},
  year={2025}
}

@article{article,
title={Cameractrl: Enabling camera control for text-to-video generation},
  author={He, Hao and Xu, Yinghao and Guo, Yuwei and Wetzstein, Gordon and Dai, Bo and Li, Hongsheng and Yang, Ceyuan},
  journal={arXiv preprint arXiv:2404.02101},
  year={2024}
}

@article{viewcrafter,
  title={Viewcrafter: Taming video diffusion models for high-fidelity novel view synthesis},
  author={Yu, Wangbo and Xing, Jinbo and Yuan, Li and Hu, Wenbo and Li, Xiaoyu and Huang, Zhipeng and Gao, Xiangjun and Wong, Tien-Tsin and Shan, Ying and Tian, Yonghong},
  journal={arXiv preprint arXiv:2409.02048},
  year={2024}
}

@inproceedings{wvd,
 title={World-consistent video diffusion with explicit 3d modeling},
  author={Zhang, Qihang and Zhai, Shuangfei and Martin, Miguel Angel Bautista and Miao, Kevin and Toshev, Alexander and Susskind, Joshua and Gu, Jiatao},
  booktitle={Proceedings of the Computer Vision and Pattern Recognition Conference},
  pages={21685--21695},
  year={2025}
}

@inproceedings{gen3c,
 title={Gen3c: 3d-informed world-consistent video generation with precise camera control},
  author={Ren, Xuanchi and Shen, Tianchang and Huang, Jiahui and Ling, Huan and Lu, Yifan and Nimier-David, Merlin and M{\"u}ller, Thomas and Keller, Alexander and Fidler, Sanja and Gao, Jun},
  booktitle={Proceedings of the IEEE/CVF Conference on Computer Vision and Pattern Recognition},
  pages={6121--6132},
  year={2025}
}

@article{geovideo,
title={Geovideo: Introducing geometric regularization into video generation model},
  author={Bai, Yunpeng and Fang, Shaoheng and Yu, Chaohui and Wang, Fan and Huang, Qixing},
  journal={Advances in Neural Information Processing Systems},
  volume={38},
  pages={57602--57622},
  year={2026}
}

@article{reconx,
title={Reconx: Reconstruct any scene from sparse views with video diffusion model},
  author={Liu, Fangfu and Sun, Wenqiang and Wang, Hanyang and Wang, Yikai and Sun, Haowen and Ye, Junliang and Zhang, Jun and Duan, Yueqi},
  journal={IEEE Transactions on Image Processing},
  year={2026},
  publisher={IEEE}
}

@inproceedings{spatia,
  title={Spatia: Video generation with updatable spatial memory},
  author={Zhao, Jinjing and Wei, Fangyun and Liu, Zhening and Zhang, Hongyang and Xu, Chang and Lu, Yan},
  booktitle={Proceedings of the IEEE/CVF Conference on Computer Vision and Pattern Recognition},
  pages={4245--4257},
  year={2026}
}

@article{fantasyworld,
  title={Fantasyworld: Geometry-consistent world modeling via unified video and 3d prediction},
  author={Dai, Yixiang and Jiang, Fan and Wang, Chiyu and Xu, Mu and Qi, Yonggang},
  journal={arXiv preprint arXiv:2509.21657},
  year={2025}
}

@article{geoflow,
  title={{GeoFlow}: Enforcing Implicit Geometric Consistency in Video Generation},
  author={Ackermann, Jan and Cai, Shengqu and Deng, Boyang and Kuang, Zhengfei and Peng, Songyou and Wetzstein, Gordon},
  journal={arXiv preprint arXiv:2605.18365},
  year={2026}
}

@article{4dnex,
  title={4dnex: Feed-forward 4d generative modeling made easy},
  author={Chen, Zhaoxi and Liu, Tianqi and Zhuo, Long and Ren, Jiawei and Tao, Zeng and Zhu, He and Hong, Fangzhou and Pan, Liang and Liu, Ziwei},
  journal={arXiv preprint arXiv:2508.13154},
  year={2025}
}

\clearpage
\appendix
\setcounter{table}{0}
\setcounter{figure}{0}
\setcounter{equation}{0}
\renewcommand{\thetable}{S\arabic{table}}
\renewcommand{\thefigure}{S\arabic{figure}}
\renewcommand{\theequation}{S\arabic{equation}}
\renewcommand{\theHtable}{supp.\arabic{table}}
\renewcommand{\theHfigure}{supp.\arabic{figure}}
\renewcommand{\theHequation}{supp.\arabic{equation}}
\section*{Appendix}

\section{Dataset Construction}
\label{sec:arxiv_supp_dataset}

Our dataset contains 1,389 image-to-video conditions: 1,339 from
COCO~\citep{mscoco} and 50 from DAVIS~\citep{davis}. We randomly hold out 100
conditions for evaluation (90 COCO and 10 DAVIS) and use the remaining 1,289
for training (1,249 COCO and 40 DAVIS).
Each condition contains a conditioning image, a subject mask on that image, an
image-to-video prompt, and a sample-specific checklist.

For DAVIS, we use the first frame of each selected video as the conditioning
image and its annotated object mask as the subject mask. For COCO, we use the
panoptic annotations to obtain subject masks. We retain motion-capable subjects,
including people, animals, and vehicles, whose masks occupy 15\%--40\% of the
image, and require an image width of at least 500 pixels. These criteria favor
scenes containing both a moving subject and sufficient static context
for geometry evaluation.

Gemini-3.1-Pro-Preview~\citep{gemini} generates a plausible continuation prompt from each
conditioning image. The prompt contains one camera-motion sentence followed by
one sentence describing the subject action and scene context. Given the image
and prompt, Gemini-3.1-Pro-Preview then constructs a sample-specific binary checklist.
All generated prompts and checklists are validated for schema compliance before
being included in the dataset.
Tab.~\ref{tab:checklist_example} shows an example checklist for one training
condition. Questions are tailored to the subject and intended motion rather
than reused across samples.

\begin{table}[!h]
  \setlength{\belowcaptionskip}{6pt}
  \caption{Example checklist for the prompt: ``Camera tracks left across the
  terrain. A brown bear walks down a dirt path in a rocky enclosure.''}
  \label{tab:checklist_example}
  \centering
  {\footnotesize
  \setlength{\tabcolsep}{4pt}
  \begin{tabularx}{\textwidth}{@{}lX@{}}
    \toprule
    Category & Binary question \\
    \midrule
    Dynamicity & Does the brown bear visibly move its body or limbs throughout
      the video, excluding apparent motion caused solely by camera panning? \\
    Physical plausibility & Does the brown bear's movement look physically
      realistic, with a natural walking gait, appropriate momentum, and proper
      foot contact with the ground? \\
    Shape & Does the brown bear maintain an anatomically plausible shape while
      allowing natural articulation, without stretching, breaking, dissolving,
      or duplicating body parts? \\
    Texture & Does the brown bear's fur color and texture remain consistent
      without unnatural flicker, pattern drift, or sudden color changes? \\
    \bottomrule
  \end{tabularx}}
\end{table}

\section{\textcolor{black}{Implementation Details}}
\label{sec:arxiv_supp_implementation}

\textcolor{black}{WorldAlign is optimized using online GRPO with LoRA adaptation
on two machines with 16 NVIDIA A800 GPUs (80 GB each). The LoRA rank and
scaling factor are 32 and 64, respectively. GRPO uses a sample group size of
$K=8$ and a clipping threshold of $\epsilon_{\mathrm{clip}}=1\times10^{-4}$.
We use AdamW with a learning rate of $1\times10^{-4}$, weight decay $10^{-4}$,
$\beta=(0.9,0.999)$, $\epsilon=10^{-8}$, and gradient-norm clipping at 1.0.
Training uses bfloat16, gradient checkpointing, and an EMA decay of 0.9
updated every four optimization steps. Each epoch samples 48 distinct
conditions and performs two gradient updates. Each method and ablation is
trained once per setting; we report checkpoints at epochs 80 and 30 for the
5B and 14B models, respectively.}

\textcolor{black}{Training rollouts contain 81 frames at $480\times832$
resolution, generated with 20 denoising steps using Flow-SDE with noise level
0.9. The guidance scales are 3.5 and 5.0 for Wan2.2-TI2V-5B and
Wan2.1-I2V-14B-480P, respectively.}

\textcolor{black}{For reward computation, SAM3~\citep{sam3} propagates the initial
subject mask through each generated video, and VGGT-1B~\citep{vggt} computes
the static and camera-motion rewards from 25 uniformly sampled frames.
The camera-motion threshold is 0.01. Qwen3.6-Plus~\citep{qwen3} evaluates the
dynamic-consistency checklist by receiving the generated video and one
question at a time and returning YES or NO. We use deterministic decoding
with temperature 0, a 10-fps video input, and at most five output tokens.
The three independently normalized reward components are equally weighted,
with $\lambda_s=\lambda_c=\lambda_d=1$.}

\textcolor{black}{During evaluation, every method generates one video for each
of the 100 held-out conditions using matched generation settings and seeds,
making all metric comparisons paired by condition. Seed 42 is used for
training, VLM decoding, and evaluation.}

\section{Human Evaluation}
\label{sec:arxiv_supp_human_evaluation}

\subsection{Protocol}

To complement the metrics in Tab.~\ref{tab:sota_comparison} of the main paper, we conduct a pairwise human evaluation on
20 conditions randomly sampled from the held-out test set. Under the
Wan2.1-I2V-14B-480P setting, WorldAlign is compared separately with the
pretrained base model, Epipolar-DPO, VideoGPA, World-R1, and GeoFlow, producing 100 unique
video pairs. Within each pair, both methods use the same conditioning input,
prompt, generation settings, and seed.
Each anonymized pair is assigned to three independent raters, yielding 300
judgments in total. Videos are shown side by side with randomized left--right
placement, and method identities are hidden. Participants choose the left
video, the right video, or a tie according to three criteria: (1) \emph{static
consistency}, whether the static environment remains geometrically coherent
throughout the video; (2) \emph{dynamic consistency}, whether the subject moves
plausibly while preserving its appearance over time; and (3) \emph{overall
preference}. We report Win/Tie/Loss percentages from the perspective of
WorldAlign.

\subsection{Results}

We summarize the evaluation at two levels. Tab.~\ref{tab:human_aggregate}
aggregates all 300 judgments to show the overall preference for WorldAlign,
whereas Tab.~\ref{tab:human_by_method} reports results against each comparator
to determine whether the trend holds across methods. All values are expressed
as Win/Tie/Loss percentages from the perspective of WorldAlign.

As shown in Tab.~\ref{tab:human_aggregate}, WorldAlign achieves win rates of
71.3\%, 75.0\%, and 76.0\% for static consistency, dynamic consistency, and
overall preference, respectively, while its loss rate remains approximately
11\% for all three criteria.

Tab.~\ref{tab:human_by_method} further shows that this advantage is not driven
by a single comparator. WorldAlign obtains majority win rates against every
method under all three criteria: 65.0\%--80.0\% for static consistency,
65.0\%--81.7\% for dynamic consistency, and 65.0\%--85.0\% for overall
preference. Together, the two tables confirm that the improvements in both
static and dynamic consistency are perceptible to human evaluators and align
with the quantitative findings in the main paper.

\begin{table}[H]
  \setlength{\belowcaptionskip}{6pt}
  \caption{Aggregate human evaluation over 300 judgments on
  Wan2.1-I2V-14B-480P. Values are Win/Tie/Loss percentages from the perspective
  of WorldAlign.}
  \label{tab:human_aggregate}
  \centering
  {\footnotesize
  \setlength{\tabcolsep}{3.4pt}
  \begin{tabular}{lccc}
    \toprule
    Criterion & Win & Tie & Loss \\
    \midrule
    Static consistency
      & 71.3 & 17.7 & 11.0 \\
    Dynamic consistency
      & 75.0 & 14.0 & 11.0 \\
    Overall preference
      & 76.0 & 12.7 & 11.3 \\
    \bottomrule
  \end{tabular}}
\par\vspace{1em}
  \setlength{\belowcaptionskip}{6pt}
  \caption{Human-evaluation results by comparator on Wan2.1-I2V-14B-480P. Each
  entry reports WorldAlign Win/Tie/Loss percentages over 60 judgments.}
  \label{tab:human_by_method}
  \centering
  {\footnotesize
  \setlength{\tabcolsep}{4.5pt}
  \begin{tabular}{lccc}
    \toprule
    Comparator & Static Win/Tie/Loss & Dynamic Win/Tie/Loss
      & Overall Win/Tie/Loss \\
    \midrule
    Base          & 73.3/11.7/15.0 & 80.0/10.0/10.0 & 81.7/6.7/11.7 \\
    Epipolar-DPO & 73.3/18.3/8.3  & 80.0/8.3/11.7  & 80.0/8.3/11.7 \\
    VideoGPA          & 65.0/26.7/8.3  & 65.0/23.3/11.7 & 68.3/23.3/8.3 \\
    World-R1           & 80.0/11.7/8.3  & 81.7/11.7/6.7  & 85.0/8.3/6.7 \\
    GeoFlow           & 65.0/20.0/15.0 & 68.3/16.7/15.0 & 65.0/16.7/18.3 \\
    \bottomrule
  \end{tabular}}
\end{table}

\section{\textcolor{black}{Complete Ablation Analysis}}
\label{sec:arxiv_supp_ablation}

\textcolor{black}{As shown in Tab.~\ref{tab:reward_ablation} of the main paper, we ablate the key components of
WorldAlign.
Starting from the `` $w/$ Masked Reprojection Reward'' in the lower half of the table, we
incrementally evaluate each component. Masked reprojection substantially reduces
M.Epi. and M.Reproj. compared with the base model, demonstrating
improved static consistency. However, Dynamic Degree simultaneously drops from
0.63 to 0.42, revealing that the low geometric errors are partially caused by
static collapse. Adding
the Camera Reward alleviates this collapse and raises Dynamic Degree to 0.70,
while both
geometric errors remain substantially below those of the base model. To further improve dynamic consistency, we add
the Dynamic Reward. Compared with ``$+$ Camera Reward'', Subject DINO increases
from 0.927 to 0.938 and the Checklist ``Yes'' Rate from 0.711 to 0.781. The two
static consistency metrics remain nearly unchanged, while Motion Smoothness
and Aesthetic Quality also improve.
To further verify the necessity of semantic masking, the upper half replaces masked reprojection with direct reprojection. Motion-induced artifacts then contaminate
the reward signals: pixels covered by these artifacts are penalized regardless of whether
the underlying static background is geometrically consistent. Consequently,
direct reprojection alone favors suppressing both camera and subject motion,
producing deceptively low geometric errors and high Subject DINO but a Dynamic
Degree of only 0.24. Adding the Camera Reward raises Dynamic Degree to 0.64, yet
the subject Dynamicity rate remains only 0.22, suggesting that the recovered
motion is dominated by the camera while the subject remains largely frozen.
Finally, adding the Dynamic Reward creates competing objectives: legitimate
subject motion improves Dynamicity and the Checklist rate, but is simultaneously
penalized by direct reprojection, weakening the effective static signal and
raising M.Epi. and M.Reproj. to 0.338 and 0.0291. In contrast, the corresponding full design in the masked-reprojection branch achieves both lower geometric errors and
substantially stronger dynamic consistency, validating the proposed decoupled
alignment.}

\section{Evaluation Details}
\label{sec:arxiv_supp_evaluation}

\subsection{Masked Epipolar Error}

Masked Epipolar Error is independent of the training reward. We sample every
15th frame at the original resolution and evaluate adjacent sampled pairs. SIFT
features are matched with a brute-force matcher and Lowe's ratio threshold 0.75.
The tracked subject mask is dilated with a $7\times7$ kernel, and matches whose
endpoints fall inside a dynamic region are removed. We estimate the fundamental
matrix with RANSAC, refit it on inliers using the normalized eight-point
algorithm, and average the square root of the Sampson distance over inlier
matches. Frame pairs with fewer than 20 valid matches are excluded.

\subsection{Masked Reprojection Error}

Masked Reprojection Error is reward-related and follows the geometry pipeline
used by the static reward. SAM3~\citep{sam3} propagates the subject mask, and
VGGT-1B~\citep{vggt} predicts point maps and cameras from 25 uniformly sampled frames. Dynamic points are
removed before constructing the shared point cloud, and dynamic pixels are
excluded from the image comparison after reprojection. We average RGB mean
squared error over valid static pixels.

\subsection{Tracked-Subject DINOv3 Similarity}

Tracked-Subject DINOv3 Similarity~\citep{dinov3} is independent of the training reward. For
every frame, we crop the square bounding region of the tracked subject, zero
pixels outside its mask, resize the crop to $224\times224$, and extract the
DINOv3 CLS feature. The metric is the mean cosine similarity between adjacent
valid subject features. Because a frozen subject can obtain a high value, we
interpret this metric jointly with Dynamicity ``Yes'' Rate and VBench Dynamic
Degree.

\end{document}